\documentclass{article}
\usepackage{amsfonts}
\usepackage{spconf,amsmath,graphicx}
\usepackage{multirow}

\title{SD-HuBERT: Sentence-Level Self-Distillation Induces \\
Syllabic Organization In HuBERT}
\makeatletter
\def\@name{
  \emph{Cheol Jun Cho}$^{1}$\qquad
  \emph{Abdelrahman Mohamed}$^{2}$\qquad
  \emph{Shang-Wen Li}$^{3}$ \\
  \emph{Alan W Black}$^{4}$ \qquad 
  \emph{Gopala K. Anumanchipalli}$^{1}$\\
}
\address{$^1$UC Berkeley \qquad $^2$ Rembrand \qquad $^3$Meta AI \qquad $^4$ Carnegie Mellon University}
\begin{document}
\ninept
\maketitle
\begin{abstract}

Data-driven unit discovery in self-supervised learning (SSL) of speech has embarked on a new era of spoken language processing. Yet, the discovered units often remain in phonetic space and speech units beyond phonemes are largely underexplored. Here, we demonstrate that a syllabic organization emerges in learning sentence-level representation of speech. In particular, we adopt ``self-distillation" objective to fine-tune the pretrained HuBERT with an aggregator token that summarizes the entire sentence. Without any supervision, the resulting model draws definite boundaries in speech, and the representations across frames exhibit salient syllabic structures. We demonstrate that this emergent structure largely corresponds to the ground truth syllables. Furthermore, we propose a new benchmark task, Spoken Speech ABX, for evaluating sentence-level representation of speech. When compared to previous models, our model outperforms in both unsupervised syllable discovery and learning sentence-level representation. Together, we demonstrate that the self-distillation of HuBERT gives rise to syllabic organization without relying on external labels or modalities, and potentially provides novel data-driven units for spoken language modeling.


\end{abstract}
\section{Introduction}

Self-supervised learning (SSL) of speech has been extremely successful in learning rich representations of speech which are transferable to many downstream tasks \cite{yang2021superb, Mohamed2022}. In particular, discrete units discovered by internal clustering of SSL models have been actively utilized for various domains, including spoken language modeling (``text-less NLP") \cite{gslm, gslm_prosody, borsos2023audiolm} and speech synthesis \cite{valle, resynthesis}. Recent studies show that speech SSL models are highly correlated with articulatory phonetics and their discretized units are fine-grained subphonemic units effectively tiling phonetic space \cite{cho2023emaprobing, hubert_unit_1, hubert_unit_2}. 

However, from a phonological viewpoint, the most naturalistic placeholder of speech is a ``syllable" rather than a phoneme. A syllable is by definition a minimal unit of pronunciation, so syllabic units are potentially better-grounded units of speech. To achieve syllabic units, a model should be able to segment the speech into a series of brackets that group phonemes. Still, the current speech SSL models significantly lack such segmentation ability.

Inspired by the success of a vision SSL model (DINO) in demonstrating the emergence of segmentation \cite{dinos}, here, we demonstrate that the same objective can induce a segmentation ability in the speech SSL model. Specifically, we fine-tune the pretrained HuBERT model with a sentence-level self-distillation method - \textbf{S}elf-\textbf{D}istilled \textbf{HuBERT} (SD-HuBERT).\footnote{Code and SSABX dataset: https://github.com/cheoljun95/sdhubert.git} Without relying on any label or external modality, SD-HuBERT naturally learns to segment continuous speech into distinct chunks, which largely correspond to the ground truth syllables. Moreover, SD-HuBERT draws salient boundaries, which allows an efficient deployment of a segmentation algorithm. 

\begin{figure}[t]
\begin{center}
\centerline{\includegraphics[width=210pt]{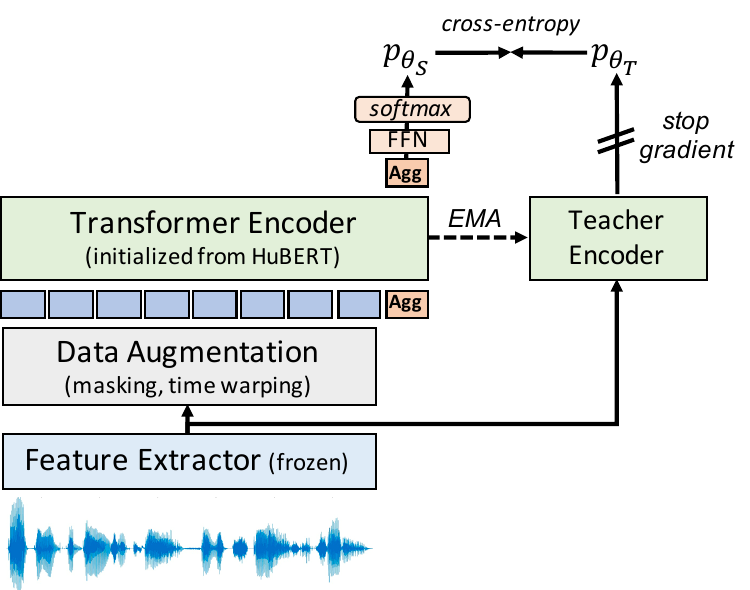}}
\vspace{-0.3cm}
\caption{Diagram of the model architecture and sentence-level self-distillation framework. An aggregator token (Agg) is inserted to summarize the entire input speech. }
\label{fig:layer}
\end{center}
\vspace{-8mm}
\end{figure}

We hypothesize that such emergent properties are driven by the enhanced representation of the model promoted by learning sentence-level information. To verify this hypothesis, we propose a new evaluation protocol, \textbf{Spoken Sentence ABX} (SSABX), for measuring the discriminability of the models on spoken sentences. This task is a tuning-free measure performed by comparing similarities between sentence-level embeddings. Our proposed model shows a higher SSABX accuracy than the baseline models including some representative speech models.

\begin{figure*}[t]
\begin{center}
\includegraphics[width=\textwidth,keepaspectratio]{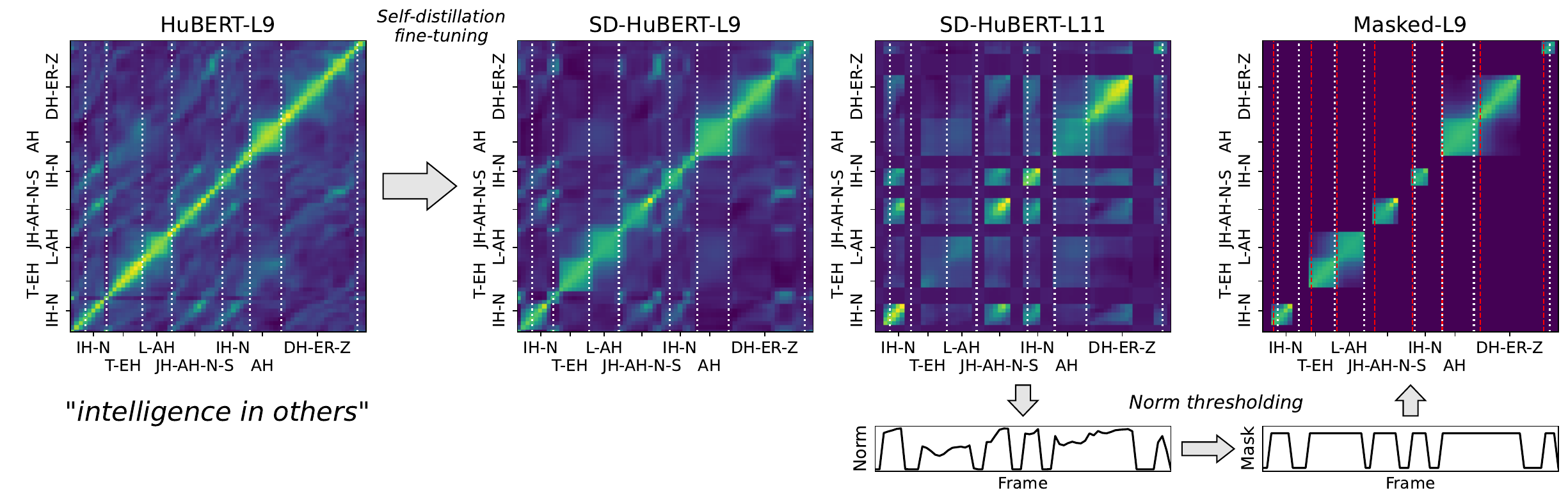}
\vspace{-0.5cm}
  \caption{Frame similarity matrices for "intelligence in others" from the 9th layer of HuBERT before fine-tuning, and the 9th and 11th layers after fine-tuning. The similarity is measured by dot product. The white dotted lines denote the ground truth syllable boundaries and the red dotted lines are the predicted boundaries. The syllables become clearly visible in SD-HuBERT after self-distillation. The frames are knocked out in the 11th layer of SD-HuBERT, drawing definite boundaries.}
  \label{fig:arch}
\end{center}
\vspace{-4mm}
\end{figure*}


Our major contributions are:
\begin{itemize}
    \item We propose a sentence-level speech representational model, SD-HuBERT, by fine-tuning pretrained HuBERT with a sentence-level self-distillation objective.
    \item We demonstrate that syllabic organization emerges in SD-HuBERT, and the model outperforms the baseline models in both syllable boundary detection and syllabic unit discovery. 
    \item SD-HuBERT infers definite sub-word boundaries by knocking out the boundary frames, which can be utilized to speed up the previous segmentation algorithm. 
    \item We propose a new benchmark task, Spoken Sentence ABX (SSABX), for simple, tuning-free evaluation of the sentence-level discriminability of speech models.
    \item When evaluated on the SSABX task, SD-HuBERT outperforms previous speech SSL models by a large margin.
\end{itemize}


\section{Related Work}

The speech processing community has long sought to discover linguistic units from speech audio. Diverse unsupervised methods have been proposed to find word boundaries and lexical semantic embeddings in speech \cite{park2007unsupervised, lee2015unsupervised, kamper2017embedded, kamper2019truly}, and recent speech SSL models have been actively leveraged to discover phonetic units from speech \cite{hsu2021hubert, van2020vector}. In particular, HuBERT \cite{hsu2021hubert} is proven to learn rich speech information and the internal clustering of the model shows high correspondence with phonemes. However, most of the previous works have been focused on lexical or phonetic units, and the unsupervised discovery of syllabic units remains underexplored. This leaves a significant gap in the transition from phonetics to higher-order linguistic components including lexicons in the speech hierarchy.

Some studies have suggested that visually grounding speech representation can reveal lexical structure in speech. They have demonstrated that words can be discovered from raw speech by learning the linkage between spoken words and their visual entities \cite{harwath2018jointly, harwath2019learning, vghubert}. 
Furthermore, Peng et al. \cite{syllable} suggest that syllabic organization emerges from visually grounded HuBERT (VG-HuBERT) \cite{vghubert} where they fine-tune HuBERT on image-spoken caption pairs to maximize shared information between the two modalities. However, we claim that cross-modal grounding is not necessary for such emergent property. Indeed, a similar emergent behavior is reported in a single-modality model in the vision domain \cite{dinos}. We empirically demonstrate this claim by fine-tuning HuBERT on the sentence-level representation of speech using speech audio data only. Given that text or image labels of speech are expensive to collect, our text-less, speech-only model can be highly beneficial.



\vspace{-4pt}

\section{Methods}
\subsection{Sentence-Level Fine-Tuning of HuBERT}

Our approach is based on a pre-trained speech SSL model, HuBERT \cite{hsu2021hubert}, which is composed of a CNN feature extractor followed by a Transformer encoder.\footnote{We use the base model with 90M parameters.} While the original model was trained for frame-level predictions, SD-HuBERT is optimized upon sentence-level representation of speech. To achieve this, an aggregator token with learnable embedding is concatenated to the inputs to the Transformer encoder \cite{vghubert, sonar_precursor, sonar}.\footnote{This token is often named [CLS] token in computer vision and NLP.} The aggregator token aggregates information across the frames into a single, representative embedding of the entire audio input. The final output of the aggregator token is passed to non-linear mapping and softmax function to parametrize the probability of the given spoken sentence. This probability is denoted as \(p_\theta(\cdot)\) where \(\theta\) represents the model weights.

We follow the self-distillation framework suggested by Caron et al. \cite{dinos}. This framework distills the student model, \(p_{\theta_S}(\cdot)\),  to the teacher model, \(p_{\theta_T}(\cdot)\), where \(\theta_T\) is an exponential moving average (EMA) of  \(\theta_S\). A random data augmentation, \(\tau(\cdot)\), is applied to the output frames of the feature extractor, which is randomly selected from a set of augmentations, \(\mathcal{T}\): random frame masking and random time warping of the frames \cite{park2019specaugment}. The masked frames are replaced with a learnable mask token. The model minimizes the cross-entropy of the probabilities inferred from the aggregator token, using the teacher inference as the target reference: \(\tau, \tau' \sim \mathcal{T}, \: \sum_{x\in X} -p_{\theta_T}(\tau(x))\:\text{log}(p_{\theta_S}(\tau'(x)))\). 
As suggested in \cite{dinos}, recentering teacher output and stop gradient are applied to prevent degeneration. 

The model weights are initialized with the weights from the official checkpoint of HuBERT, which is trained on 960 hours of English speech from LibriSpeech data \cite{panayotov2015librispeech}. We reinitialize the last three layers of the Transformer encoder with random weights. 
As the data augmentation is applied after the feature extractor, we freeze the feature extractor and the positional encoding model.\footnote{Otherwise, the model adapts to the data augmentation and degenerates.}

We use LibriSpeech data for training and evaluating, which is exactly the same as the original HuBERT training. Five-second windows are randomly sampled from each audio clip to reduce computational complexity. AdamW \cite{adamw} is utilized for the optimizer with a batch size of 100 for 200K iterations. The learning rate starts with 1e-4 and decays to 1e-5 by the Cosine learning rate schedule. For EMA of the teacher model, we set the decay rate as 0.999.

\subsection{Unsupervised Syllable Discovery}

The proposed self-distillation shapes the embedding space with interesting topology as shown in the frame similarity matrices in Fig. \ref{fig:arch}. While the similarities are relatively local in the original HuBERT (Fig. \ref{fig:arch}, HuBERT-L9), after the self-distillation, the similarities span longer windows, largely overlapping with the ground truth syllables (Fig. \ref{fig:arch}, SD-HuBERT-L9). 
Moreover, in the later layers of SD-HuBERT, some definite boundaries are drawn (Fig. \ref{fig:arch}, SD-HuBERT-L11). Frames near the boundaries are knocked out to have distinctively small norms. This phenomenon happens in the last randomly initialized layers, which is most salient in the 11th layer. This is not observable when we remove the random reinitialization. Leveraging such indicators, the input speech can be easily segmented by thresholding the frame norm with a constant value (``Norm thresholding" in Fig. \ref{fig:arch}). However, the resulting segments are not yet syllables. In the example of speaking ``intelligence in others" (Fig. \ref{fig:arch} Masked-L9), the norm thresholding assigns a single segment for ``T-EH" and ``L-AH". The same issue happens with ``AH" and ``DH-ER-Z" in the later frames. Therefore, these segments may span more than one syllable, thus we applied the minimum cut algorithm \cite{syllable} to refine each segment.\footnote{In general, the number of syllables per segment is not bigger than three.}


This novel emergent behavior of SD-HuBERT provides a first cut of segmentation for free, reducing the search space of the mincut algorithm by a large margin. The original method has \(O(kN^2)\) time complexity where \(k\), \(N\) is the number of syllables and frames, respectively. As the norm thresholding divides the frames by the number of syllables asymptotically, our model can reduce time complexity down to \(O(N^2/k)\). 

Other than the segmentation algorithm, the rest of the procedure largely follows Peng et al. \cite{syllable}. To evaluate the detected syllable boundaries, we measured precision (Pr), recall (Re), F1, and R scores with the 50 ms tolerance window. Although the ground truth syllable boundaries are seamlessly annotated, the predicted boundaries are not due to the knocked-out frames. Therefore, we use the onsets of the segments as the detected boundaries. 

In addition to evaluating the segmentation, we apply clustering analysis to measure how well the segments correspond to the ground truth syllables. The features within segments are averaged to be segment-wise features and then clustered to form a set of data-driven syllabic units. We apply two steps of clustering by initially assigning a large number of clusters (\# = 16384) and then merging clusters by agglomerative clustering on the cluster centers (\# = 16384 \(\rightarrow\) 4096). Then, following \cite{hsu2021hubert, syllable}, we measure purity terms, syllable purity (SP) and cluster purity (CP), which measures how purely a unit category is mapped to a most matching syllable (SP) and vice versa (AP). The Hungarian matching algorithm is leveraged to match unit categories to ground truth syllables, maximizing the intersection-over-union between matching unit segments and labeled syllable spans. The test split of LibriSpeech is used for the evaluation where the ground truth labels are obtained by Montreal Forced Alignment \cite{mfa} and the syllabification of the transcribed texts.

\subsection{Spoken Sentence ABX (SSABX)}
Inspired by Semantic Textual Similarity (STS) tasks in NLP \cite{agirre2016semeval},  we propose a new benchmark task by carefully mining triplets from the LibriSpeech test set. Unlike the multi-categorical rating in STS, we design an ABX task focused on the sentence discriminability of speech models. 
First of all, each audio in the LibriSpeech test set is segmented into smaller pieces of sentences by cutting silent moments. Then, we leverage an off-the-shelf textual sentence embedding model, SimCSE \cite{gao2021simcse}, to extract the ground truth sentence embedding of the transcribed texts. The similarity between two sentences is measured by cosine similarity of the inferred sentence embeddings, and a pair with higher similarity is regarded as the positive pair in an ABX triplet. We carefully designed the following criteria for curating the test set of \((X, Pos, Neg)\) triplets.
\begin{itemize}
\itemsep0em 
\item The matching condition of the positive pair has cosine similarity higher than or equal to 0.8.
\item To balance the difficulty of the ABX task, the range of similarity of negative samples is divided into three groups, [-1, 0.2], [0.2, 0.4], [0.4, 0.6], and 1K samples are sampled for each group.
\item The difference in the number of words between \(X\) and \(Pos\), and \(X\) and \(Neg\) is less than four words.
\item Every speech in the triplet is from different speakers.
\item To prevent making decisions based on acoustic or phonetic similarity, we rejected samples with a high Levenshtein similarity ratio (\(>\) 0.7) on the text between \(X\) and \(Pos\).
\item Each sentence has at least five words and the speech does not exceed five seconds.
\end{itemize}
The final test set includes 3K triplets of spoken sentences.
An example triplet is:\\
--  \(X\): \textit{“and must have locked the door when you went out”}\\
--  \(Pos\): \textit{``She found the door but it was locked outside”}\\
--  \(Neg\): \textit{``and the horse a going like a house afire too”}\\

\begin{table}[t]
\caption{Performance of syllable boundary detection and clustering by different models [\%]. TC is time complexity.}
\begin{center}
\label{tab:syl_results}
\begin{tabular}{c|c|c|c|c|c|c|c}
\hline
Model                           & Pr          & Re          & F1          & R           & SP          & CP          & TC                        \\ \hline
HuBERT                          & 47          & 27          & 35          & 47          & 28          & 30          & \(O(kN^2)\)  \\
VG-HuBERT                       & 63          & 64          & 64          & 69          & 53          & 43 & \(O(kN^2)\)  \\ \hline
SD-HuBERT                       & \textbf{64} & \textbf{71} & \textbf{67} & \textbf{71} & \textbf{54} & \textbf{46}          & \(O(N^2/k)\) \\
\multicolumn{1}{r|}{-- mincut} & \textbf{69}          & 58          & 63          & 68          & 38          & 43          & \(O(N)\)                      \\ \hline
\end{tabular}
\end{center}
\vspace{-4mm}
\end{table}

\section{Results}
\subsection{Evaluation on Syllable Boundaries and Clustering}

We compare the proposed SD-HuBERT with HuBERT and VG-HuBERT. The 11th layer of SD-HuBERT is used for norm thresholding, while the 9th layer is employed for the minimum cut algorithm and clustering. For HuBERT, we use the 9th layer and for VG-HuBERT, we followed the exact configuration using the checkpoint released by the authors.\footnote{https://github.com/jasonppy/syllable-discovery.git} Table \ref{tab:syl_results} compares the syllable boundaries and clustering scores by SD-HuBERT and the baseline models. As shown in the table, SD-HuBERT outperforms the baselines in all evaluation metrics. Furthermore, the time complexity of the segmentation in the proposed method is significantly faster than that of the baselines. For a typical sentence with 25-30 syllables, our method can boost up to several hundred times compared to the previous method. This is even faster without the minimum cut algorithm with the time complexity of \(O(N)\), which provides a higher precision score with some compromise on other metrics. The overall results suggest that SD-HuBERT can more effectively and efficiently discover syllabic units from speech compared to the baselines.

\subsection{Evaluation on Sentence-level Speech Embedding}
We evaluated some representative speech SSL models \cite{baevski2020wav2vec,hsu2021hubert, chen2022wavlm}, VG-HuBERT and a text-based word embedding, GloVe \cite{pennington2014glove}, along with variations of our model.\footnote{We used base models for Wav2Vec2 and HuBERT, and the large model for WavLM, the current SOTA speech SSL model.} The sentence-level embeddings are extracted from the models by averaging the frame-wise embeddings within sentences or from the aggregator tokens if applicable. We test every layer in the models and the scores from the best layers are reported in Table \ref{tab:sem_results}.

\begin{table}[t]
\caption{Accuracy [\%] of the SSABX task by different models, using frame average (Favg) or aggregator token (Agg).}
\vspace{+3mm}
\begin{center}
\label{tab:sem_results}

\begin{tabular}{c|l|cc}
\hline
\multirow{2}{*}{Modality} & \multicolumn{1}{c|}{\multirow{2}{*}{Model}} & \multicolumn{2}{c}{Acc [\%] }                \\ \cline{3-4} 
                          & \multicolumn{1}{c|}{}                       & \multicolumn{1}{c|}{Agg} & Favg        \\ \hline
\multirow{2}{*}{Text}     & \multicolumn{1}{c|}{SimCSE}                                     & \multicolumn{1}{c|}{100} & --          \\
                          & \multicolumn{1}{c|}{GloVe \cite{pennington2014glove}}   & \multicolumn{1}{c|}{--}  & 97          \\ \hline
\multirow{8}{*}{Speech}   & \multicolumn{1}{c|}{Wav2Vec2 \cite{baevski2020wav2vec}} & \multicolumn{1}{c|}{--}  & 74          \\
                          & \multicolumn{1}{c|}{HuBERT}                                      & \multicolumn{1}{c|}{--}  & 84          \\
                          & \multicolumn{1}{c|}{WavLM \cite{chen2022wavlm}} & \multicolumn{1}{c|}{--}  & 87          \\
                          & \multicolumn{1}{c|}{VG-HuBERT}                                   & \multicolumn{1}{c|}{\textbf{77}}  & 72          \\ \cline{2-4} 
                          & \multicolumn{1}{c|}{SD-HuBERT}                                   & \multicolumn{1}{c|}{63}  & \textbf{90} \\
                          & \multicolumn{1}{r|}{-- re-init}                                  & \multicolumn{1}{c|}{53}  & \textbf{91} \\
                          & \multicolumn{1}{r|}{+ all-re-init}                               & \multicolumn{1}{c|}{46}  & 46          \\ \hline
\end{tabular}
\end{center}
\vspace{-4mm}
\end{table}

When compared to other speech models, SD-HuBERT outperforms by a large margin, achieving 90\% accuracy with frame averaging (Favg). The accuracy significantly drops in VG-HuBERT, the HuBERT fine-tuned on image-speech pairs, indicating that visual grounding may harm the speech representation. One potential reason is that the visual grounding limits the coverage of speech because not all spoken terms have visual entities; for example, abstract words like \textit{``love"}. Without reinitializing the last three layers, the model shows a similar score (``-- re-init" in Table \ref{tab:sem_results}). However, the model fails severely with all Transformer layers initialized randomly, showing a score even below the chance level (``+ all-re-init" in Table \ref{tab:sem_results}). This suggests that the initial starting point as pretrained HuBERT is critical to train the model properly.

However, using the representation directly from the aggregator token (Agg) is significantly worse than using Favg, and it is even worse without the last layer initialization. This indicates that the information in the aggregator might be dominated by paralinguistic information rather than linguistic content, which requires more analyses to fully grasp the characteristics of this aggregator token. On the other hand, the aggregator token shows better performance than the frame average in VG-HuBERT, where some paralinguistic factors would be marginalized by visual grounding. 

Among the other baselines, HuBERT outperforms Wav2Vec2. Since those two models only differ in training objective, the SSABX performance may be significantly influenced by the training objective. WavLM, the current state-of-the-art speech model, achieves the highest score among speech models. Lastly, the gap from simple word embedding (GloVe) suggests potential room for improvement.

\subsection{Why does syllabic organization emerge in SD-HuBERT?}

In common speech SSL approaches including HuBERT, the model output is factorized by each frame, and representation learning is empowered by predicting randomly masked frames. To accomplish the masked prediction, the model preferably learns the local dynamics across frames as shown in Fig. \ref{fig:arch}, which is supported by a probing study against dynamical articulatory features \cite{cho2023emaprobing}. However, with the absence of frame-level prediction, the model may make a more parsimonious choice for representing speech, which ends up marginalizing local articulatory dynamics. Indeed, the layer-wise analysis reveals that the articulatory information diminishes in the later layer after the fine-tuning, while the SSABX score increases (Fig. \ref{fig:layer}). Though a more extensive analysis is required to verify this hypothesis, our work made an important step toward revealing how speech can be naturally segmented without any supervision and a natural selection of such segmentation is a syllable.



\begin{figure}[t]
\begin{center}
\centerline{\includegraphics[width=170pt]{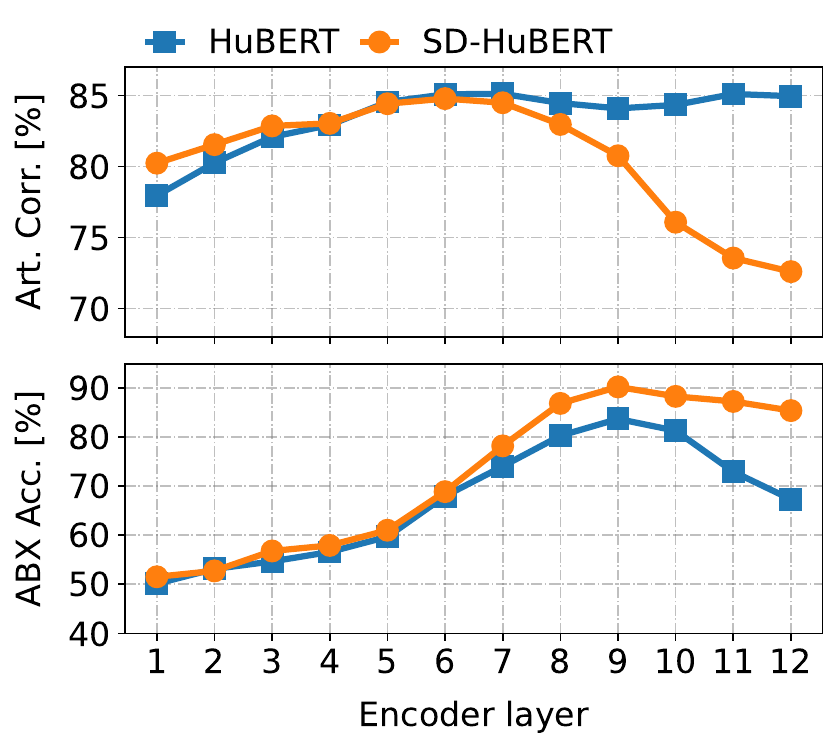}}
\vspace{-0.3cm}
\caption{Layer-wise analysis of articulatory correlation \cite{cho2023emaprobing} (top) and SSABX performance (bottom) of HuBERT (blue) and SD-HuBERT (orange).}
\label{fig:layer}
\end{center}
\vspace{-8mm}
\end{figure}

\section{Conclusion}

By fine-tuning HuBERT with sentence-level self-distillation, a syllabic organization emerges without any supervision or relying on cross-modal reference. The data-driven discovery of syllables offered by our model is more effective and efficient than the previous approaches. As syllables are phonologically grounded units of speech, our novel syllabic units discovered by SD-HuBERT may serve as an effective interface for spoken language models and various speech downstream tasks. 


\section{Acknowledgements}

This research is supported by the following grants to PI Anumanchipalli --- NSF award 2106928, BAIR Commons-Meta AI Research, the Rose Hills Innovator Program, and UC Noyce Initiative, at UC Berkeley.



\label{sec:refs}
\bibliographystyle{IEEEbib}
\fontsize{9}{9}\selectfont
\bibliography{strings,refs}

\end{document}